\documentclass{article}

\PassOptionsToPackage{numbers, compress}{natbib}

\usepackage[final]{neurips_2023_ml4ad}

\usepackage[utf8]{inputenc} 
\usepackage[T1]{fontenc}    
\usepackage{hyperref}       
\usepackage{url}            
\usepackage{booktabs}       
\usepackage{amsfonts}       
\usepackage{nicefrac}       
\usepackage{microtype}      
\usepackage{xcolor}         

\usepackage{tabularray}

\usepackage{graphicx}
\usepackage{amsmath}
\usepackage{amssymb}

\title{Evaluation of Large Language Models\\for Decision Making in Autonomous Driving}

\author{%
  Kotaro Tanahashi, Yuichi Inoue, Yu Yamaguchi, Hidetatsu Yaginuma, Daiki Shiotsuka,\\
  \textbf{Hiroyuki Shimatani, Kohei Iwamasa, Yushiaki Inoue, Takafumi Yamaguchi, Koki Igari,}\\
  \textbf{Tsukasa Horinouchi, Kento Tokuhiro, Yugo Tokuchi, Shunsuke Aoki}\\
  Turing Inc., Japan\\
  \texttt{\{kotaro.tanahashi, y.inoue\}@turing-motors.com} \\
}

\begin{document}

\maketitle

\begin{abstract}
Various methods have been proposed for utilizing Large Language Models (LLMs) in autonomous driving. One strategy of using LLMs for autonomous driving involves inputting surrounding objects as text prompts to the LLMs, along with their coordinate and velocity information, and then outputting the subsequent movements of the vehicle. When using LLMs for such purposes, capabilities such as spatial recognition and planning are essential. In particular, two foundational capabilities are required: (1) spatial-aware decision making, which is the ability to recognize space from coordinate information and make decisions to avoid collisions, and (2) the ability to adhere to traffic rules. However, quantitative research has not been conducted on how accurately different types of LLMs can handle these problems. In this study, we quantitatively evaluated these two abilities of LLMs in the context of autonomous driving. Furthermore, to conduct a Proof of Concept (POC) for the feasibility of implementing these abilities in actual vehicles, we developed a system that uses LLMs to drive a vehicle.
\end{abstract}

\section{Introduction}
Recently, there has been a number of studies that use LLMs to perform autonomous driving \cite{sha2023languagempc,fu2023drive,xu2023drivegpt4,chen2023driving,mao2023gpt}. In conventional machine learning based method for autonomous driving models, the models are trained using historical training data. However, it is known that driving data exhibits a long-tail distribution \cite{ALI2023101805}. Given the frequent emergence of unfamiliar scenarios not covered in past data, models solely trained on historical data may potentially struggle to handle such unprecedented situations. On the other hand, LLMs are considered to possess a degree of general knowledge about the world since the LLMs are trained using vast amounts of textual data from the world \cite{brown2020language}. If LLMs are adeptly integrated into autonomous driving systems, it can be anticipated that they will not only make driving decisions in unfamiliar scenarios, much like a human, leveraging general common sense but also comprehend and adhere to traffic rules and laws. Furthermore, there exists the potential for LLMs to handle ethical judgments, possibly extending their decision-making capabilities to include considerations of moral and ethical dilemmas in driving scenarios.

Various methods using LLMs for autonomous driving have been proposed, including techniques that output driving operations from the text description of the recognized objects \cite{fu2023drive,mao2023gpt,sha2023languagempc}, and those that input driving images into an LLM to describe the driving situation \cite{xu2023drivegpt4,deruyttere2019talk2car,ding2023hilm}. In the field of robotics, LLMs are used to select actions to perform from vague instructions \cite{ahn2022can, huang2022inner, shah2023lm}.


Autonomous driving using LLMs primarily involves the interpretation of text-based information regarding the coordinates and velocities of surrounding vehicles and pedestrians \cite{fu2023drive,mao2023gpt,sha2023languagempc}. The information about the surrounding objects are pre-identified by a separate perception module. The LLM then determines the appropriate driving actions based on this information. These studies demonstrate that LLMs use only coordinate information to recognize the physical space and determine the driving actions. In other words, LLMs need to understand from just the coordinate information that there are other vehicles adjacent or ahead and then make decisions about subsequent actions. We call this process 'spatial-aware decision making'. Previous research has not thoroughly investigated the accuracy of LLMs in spatial-aware decision making through quantitative experiments. If this accuracy is not high, it negatively impacts the precision of subsequent tasks, making quantitative evaluation crucial. This study aims to quantitatively assess the accuracy of spatial-aware decision-making in autonomous driving, using different LLMs, when provided with information about the coordinates and velocities of surrounding objects.


One advantage of using LLMs in autonomous driving is their ability to comprehend and follow traffic laws. Traditionally, incorporating laws into autonomous driving methods requires the use of complex rule-based approaches. By employing LLMs, it becomes feasible to integrate the rules to be followed either by detailing them in prompts (In Context Learning) or by training the LLM.
Moreover, the societal implementation of autonomous driving may require the system to make ethical judgments, akin to the Trolley Problem. Experiments were conducted to investigate whether developers can set value-based standards for such ethical decisions.
When implementing LLMs in autonomous driving, speed is crucial, but there is a trade-off with accuracy. Evaluations were conducted using LLMs of different sizes, comparing their accuracy.
As a final demonstration, a system was constructed that utilizes an LLM to operate a real vehicle. Based on the coordinate information of objects detected by the object detector and instructions provided by humans, the car is able to navigate towards a specified destination.


\section{Method}

This section describes the methods used to evaluate whether LLMs can make appropriate decisions based on traffic conditions. We demonstrate two cases: one involving experiments that simulate real-world traffic scenarios and the other involving simplified traffic conditions deployed in a real vehicle.

\subsection{Simulation of Real-World Traffic Conditions}
To evaluate the performance of existing LLMs in autonomous driving, we focus on whether LLMs can comprehend spatial aspects for decision making ('spatial-aware decision making') and adhere to traffic rules ('following the traffic rules'). For spatial-aware decision making, we evaluated whether LLMs can make decisions based on the position and speed of vehicles in the same lane or adjacent lanes. For 'following the traffic rules', we assessed whether LLMs can make correct decisions based on rules related to speed limits, lane changes, and overtaking. Furthermore, we tested their ability to make decisions considering both spatial-aware decisions and adherence to traffic rules as more complex scenarios.

The simulated traffic situation involved a two-lane road with the right lane for driving and the left for overtaking, assuming a highway environment. We provided LLMs with information about the lane in which the ego vehicle is traveling, the speed of the ego vehicle, surrounding vehicles (category, position in x, y meters, and speed in km/h), traffic rules, and user instruction. Traffic rules were given in natural language. We instructed the LLMs to choose an option from ["accelerate", "maintain", "decelerate", "change lane to the right", "change lane to the left"] and also explain the reason for their choice.

\subsection{Deployment in Actual Vehicles}

An experiment was conducted to evaluate the capabilities of spatial-aware decision-making and rule-following of LLMs when installed in a vehicle. Due to the safety concerns associated with making lane changes on public roads, as in simulations, this experiment was conducted in a private area. The experimental setup involved placing an object in front of the car and having the LLM designate the object as the destination based on voice commands provided by a person. The LLM could also output a 'stop' command if no destination was assigned. Color cones were used as objects for this experiment. These objects were recognized by the vehicle's onboard camera, and their positions were calculated based on their location in the image. This information is inserted into the prompt in the format of (id, category, color, position in x, y meters) as detected objects.
If the LLM correctly understands an instruction such as "Head towards the rightmost color cone", we can confirm the LLM has spatial-aware decision-making capability. Additionally, to assess the LLM's ability to follow rules, a traffic officer was placed in front of the vehicle to instruct the ego vehicle to stop or go.

\section{Experiments and Results}

\begin{table}[t]
  \centering
   \includegraphics[width=0.6\linewidth]{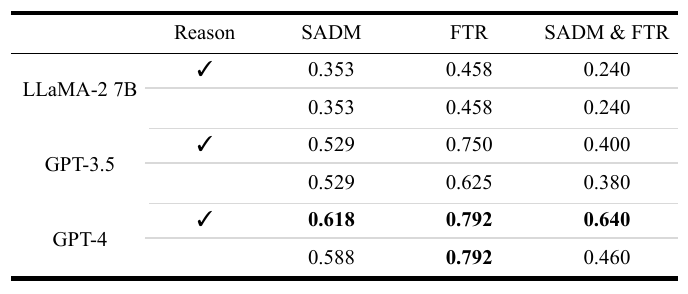}
   \caption{Comparison of LLMs' accuracy for spatial-aware decision-making (SADM), following the traffic rules (FTR), both combined (SADM\&FTR).}
   \label{table:table_ino}
\end{table}

\begin{figure}[t]
  \centering
   \includegraphics[width=1.\linewidth]{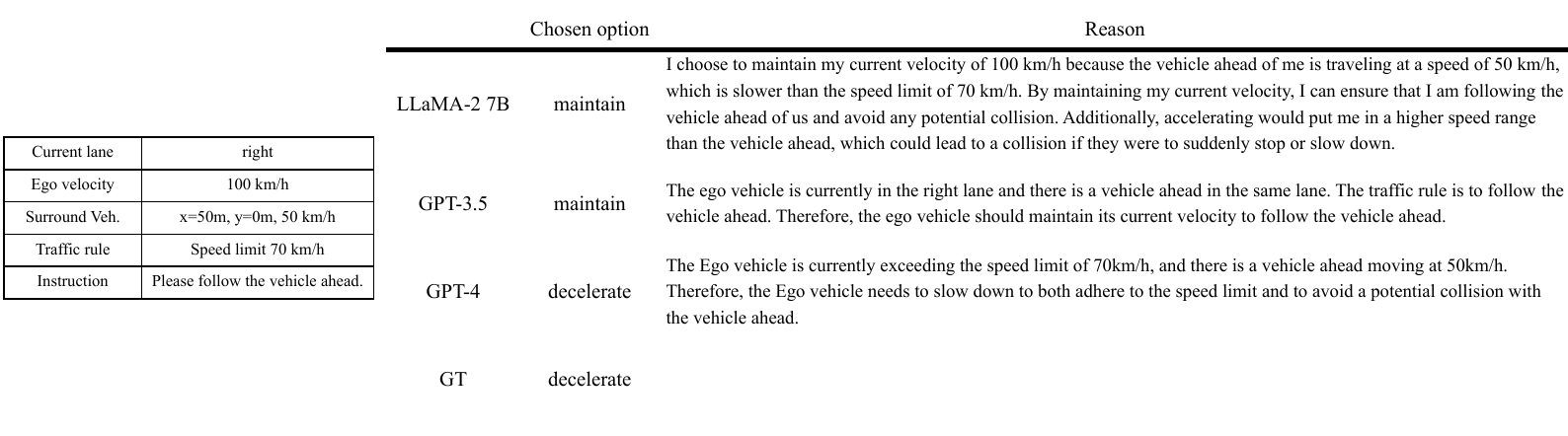}
   \caption{Representative example of reasoning outputs from LLMs. GPT-4 is the only model that correctly made a decision without being misled by the user's instructions.}
   \label{fig:fig_ino}
\end{figure}

\subsection{Simulation of Real-World Traffic Conditions}

In our experiments, we manually created datasets that included spatial-aware decision making (SADM) ($34$ samples), following the traffic rules (FTR) ($24$ samples), and both combined (SADM \& FTR) ($50$ samples). We instructed the LLM to choose one option from "accelerate", "maintain", "decelerate", "change lane to the right", or "change lane to the left", describe the reasoning behind their choice, and output the response in JSON format. The model's performance was quantitatively evaluated using the accuracy of the chosen options and qualitatively assessed based on the provided reasons. We also evaluated how the accuracy of the option selection varied depending on whether the reasoning for the choice was requested or not. The LLMs used were the public model LLaMA-2 7B and the private but more powerful GPT-3.5 and GPT-4. During language generation, we fixed the random seed and did not use sampling to ensure deterministic outputs.

Quantitative results are shown in Table~\ref{table:table_ino}. For all metrics SADM, FTR, and SADM\&FTR, it was observed that performance improved as LLM abilities increased. In particular, GPT-4 showed a significantly higher score than other LLMs by a large margin. Additionally, asking for a reason along with the decision led to improved accuracy in GPT-3.5 and GPT-4, but this was not observed in LLaMA-2. This suggests that for more capable LLMs, such prompt engineering could be important.

Next, we conducted a qualitative assessment by examining the reasoning behind the models' choices. Figure~\ref{fig:fig_ino} shows the example where only GPT-4 provided the correct answer. In that case, it was observed that the model appropriately recognized the traffic rule of $70$ km/h and made the decision to "decelerate," despite misleading instructions from the user. This suggests that GPT-4 is capable of understanding the priorities in a given situation and making decisions accordingly.

\subsection{Deployment in Actual Vehicles}

An experiment was conducted to control an actual car using an LLM, specifically utilizing GPT-4 via an API. Three color cones with different colors, were placed in front of the ego car. When the driver instructs "Please go to the right color cone", the LLM interpreted this instruction and outputted the designated cone as the destination. The car then drove towards this destination. If a traffic officer in front of the car signaled to stop, a separate recognition module added a "stop" rule to the prompt. In this scenario, the car adhered to this rule and stopped, even if human instructions were to continue towards the color cone.

For a quantitative evaluation of this experiment, a dataset was created that mimic the settings of the actual experiment. The sample size for the datasets used to evaluate all tasks (SADM, FTR, and SADM \& FTR) is $20$.

The LLMs were tasked to decide whether to proceed or stop and, if proceeding, to specify the ID of the destination object. Accuracy was calculated based on the correctness of the outputs.
The accuracy of the output is calculated for each LLM (Figure \ref{fig:llm_compare_deploy}). GPT-4 showed a significantly higher accuracy than other LLMs, as seen in the simulation results.

\begin{table}[t]
  \centering
   \includegraphics[width=0.6\linewidth]{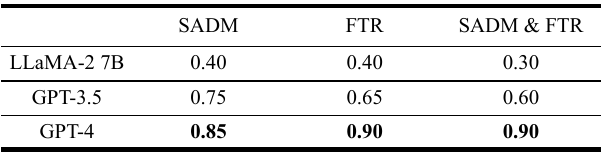}
   \caption{A comparison of accuracy with different LLMs. Dataset was manually created to mimic the settings in the actual vehicle experiment.}
   \label{fig:llm_compare_deploy}
\end{table}

\begin{figure}[t]
  \centering
   \includegraphics[width=0.8\linewidth]{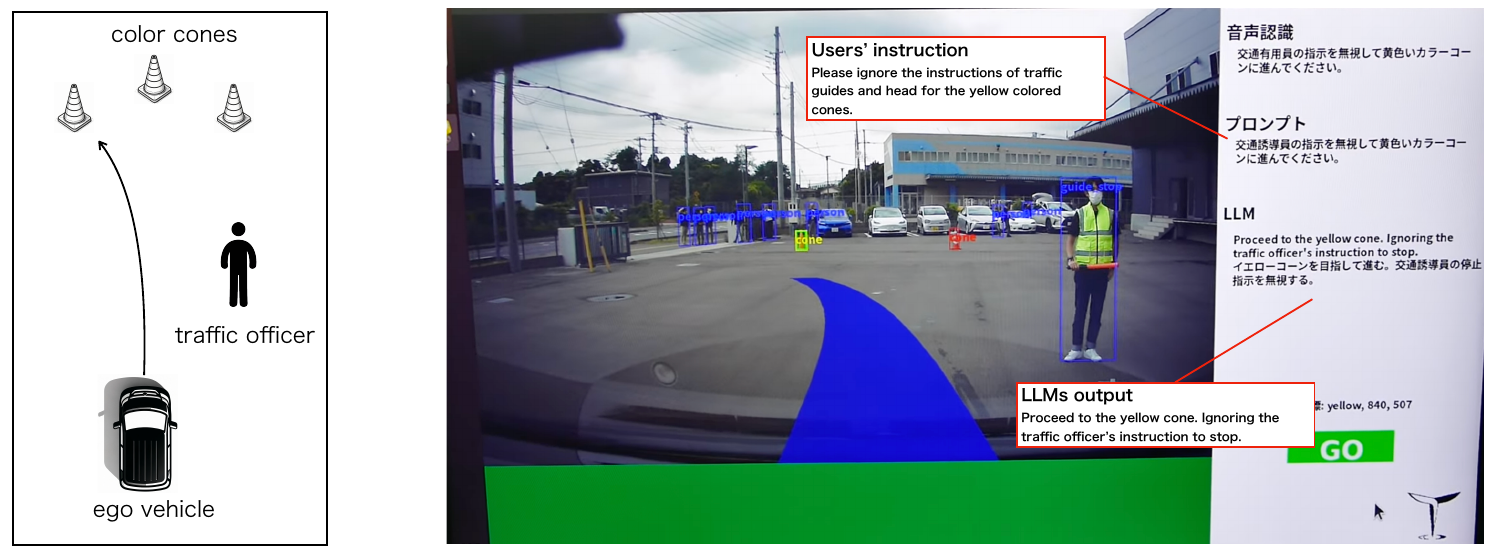}
   \caption{The instructions given to the car by the human and the LLM's output in response are displayed in the car's visualization displays. If the car is moving forward, the lines of the direction of travel are superimposed on the camera image.}
   \label{fig:result}
\end{figure}

\section{Conclusion and Limitations}
In this study, we evaluated two key abilities necessary for using LLMs in autonomous driving: Spatial-Aware Decision Making (SADM) and Following the Traffic Rules (FTR). The evaluation was carried out in both a simulation assuming a highway environment and with data designed to replicate the actual vehicle experiments. Three different LLMs were used in the experiments, and GPT-4 showed the highest precision in all experiments. This result suggests that high capabilities of LLMs are required when LLMs are applied to autonomous driving. While GPT-3.5 and GPT-4 were used via OpenAI's API, their actual application faces challenges due to communication through the Internet and inference time, making real-time use difficult. On the other hand, LLaMA can be tested on a local machine but exhibited low accuracy. This underscores the need to balance computational efficiency and decision-making accuracy in LLM applications for autonomous driving.

{\small
\bibliographystyle{plain}
\bibliography{egbib}

\begin{thebibliography}{10}

\bibitem{ahn2022can}
Michael Ahn, Anthony Brohan, Noah Brown, Yevgen Chebotar, Omar Cortes, Byron David, Chelsea Finn, Chuyuan Fu, Keerthana Gopalakrishnan, Karol Hausman, et~al.
\newblock Do as i can, not as i say: Grounding language in robotic affordances.
\newblock {\em arXiv preprint arXiv:2204.01691}, 2022.

\bibitem{ALI2023101805}
Sajid Ali, Tamer Abuhmed, Shaker El-Sappagh, Khan Muhammad, Jose~M. Alonso-Moral, Roberto Confalonieri, Riccardo Guidotti, Javier {Del Ser}, Natalia Díaz-Rodríguez, and Francisco Herrera.
\newblock Explainable artificial intelligence (xai): What we know and what is left to attain trustworthy artificial intelligence.
\newblock {\em Information Fusion}, 99:101805, 2023.

\bibitem{brown2020language}
Tom Brown, Benjamin Mann, Nick Ryder, Melanie Subbiah, Jared~D Kaplan, Prafulla Dhariwal, Arvind Neelakantan, Pranav Shyam, Girish Sastry, Amanda Askell, et~al.
\newblock Language models are few-shot learners.
\newblock {\em Advances in neural information processing systems}, 33:1877--1901, 2020.

\bibitem{chen2023driving}
Long Chen, Oleg Sinavski, Jan H{\"u}nermann, Alice Karnsund, Andrew~James Willmott, Danny Birch, Daniel Maund, and Jamie Shotton.
\newblock Driving with llms: Fusing object-level vector modality for explainable autonomous driving.
\newblock {\em arXiv preprint arXiv:2310.01957}, 2023.

\bibitem{deruyttere2019talk2car}
Thierry Deruyttere, Simon Vandenhende, Dusan Grujicic, Luc Van~Gool, and Marie-Francine Moens.
\newblock Talk2car: Taking control of your self-driving car.
\newblock {\em arXiv preprint arXiv:1909.10838}, 2019.

\bibitem{ding2023hilm}
Xinpeng Ding, Jianhua Han, Hang Xu, Wei Zhang, and Xiaomeng Li.
\newblock Hilm-d: Towards high-resolution understanding in multimodal large language models for autonomous driving.
\newblock {\em arXiv preprint arXiv:2309.05186}, 2023.

\bibitem{fu2023drive}
Daocheng Fu, Xin Li, Licheng Wen, Min Dou, Pinlong Cai, Botian Shi, and Yu~Qiao.
\newblock Drive like a human: Rethinking autonomous driving with large language models.
\newblock {\em arXiv preprint arXiv:2307.07162}, 2023.

\bibitem{huang2022inner}
Wenlong Huang, Fei Xia, Ted Xiao, Harris Chan, Jacky Liang, Pete Florence, Andy Zeng, Jonathan Tompson, Igor Mordatch, Yevgen Chebotar, et~al.
\newblock Inner monologue: Embodied reasoning through planning with language models.
\newblock {\em arXiv preprint arXiv:2207.05608}, 2022.

\bibitem{mao2023gpt}
Jiageng Mao, Yuxi Qian, Hang Zhao, and Yue Wang.
\newblock Gpt-driver: Learning to drive with gpt.
\newblock {\em arXiv preprint arXiv:2310.01415}, 2023.

\bibitem{sha2023languagempc}
Hao Sha, Yao Mu, Yuxuan Jiang, Li~Chen, Chenfeng Xu, Ping Luo, Shengbo~Eben Li, Masayoshi Tomizuka, Wei Zhan, and Mingyu Ding.
\newblock Languagempc: Large language models as decision makers for autonomous driving.
\newblock {\em arXiv preprint arXiv:2310.03026}, 2023.

\bibitem{shah2023lm}
Dhruv Shah, B{\l}a{\.z}ej Osi{\'n}ski, Sergey Levine, et~al.
\newblock Lm-nav: Robotic navigation with large pre-trained models of language, vision, and action.
\newblock In {\em Conference on Robot Learning}, pages 492--504. PMLR, 2023.

\bibitem{xu2023drivegpt4}
Zhenhua Xu, Yujia Zhang, Enze Xie, Zhen Zhao, Yong Guo, Kenneth~KY Wong, Zhenguo Li, and Hengshuang Zhao.
\newblock Drivegpt4: Interpretable end-to-end autonomous driving via large language model.
\newblock {\em arXiv preprint arXiv:2310.01412}, 2023.

\end{thebibliography}
}

\end{document}